\documentclass[10pt,twocolumn]{article}

\usepackage[utf8]{inputenc}
\usepackage[T1]{fontenc}
\usepackage{lmodern}

\usepackage[margin=1in]{geometry}

\usepackage{amsmath}
\usepackage{amssymb}
\usepackage{amsthm}

\usepackage{graphicx}
\usepackage{float}
\usepackage[section]{placeins}  
\usepackage{subcaption}
\usepackage{capt-of}  
\graphicspath{{figures/}}

\usepackage{booktabs}
\usepackage{multirow}

\usepackage{algorithm}
\usepackage{algpseudocode}

\usepackage{listings}
\usepackage[hidelinks]{hyperref}
\usepackage{cleveref}

\title{%
    \textbf{Aeon}: High-Performance Neuro-Symbolic Memory Management \\
    for Long-Horizon LLM Agents%
}

\author{%
    Mustafa Arslan \\
    \textit{Independent Researcher, Istanbul, Turkey}
}

\date{\vspace{-4ex}}

\begin{document}

\maketitle

\begin{abstract}
Large Language Models (LLMs) are fundamentally constrained by the quadratic computational cost of self-attention and the ``Lost in the Middle'' phenomenon, where reasoning capabilities degrade as context windows expand. Existing solutions, primarily ``Flat RAG'' architectures relying on vector databases, treat memory as an unstructured bag of embeddings, failing to capture the hierarchical and temporal structure of long-horizon interactions. This paper presents \textbf{Aeon}, a Neuro-Symbolic Cognitive Operating System that redefines memory as a managed OS resource. Aeon structures memory into a \textbf{Memory Palace} (a spatial index implemented via \textsc{Atlas}, a SIMD-accelerated \textbf{Page-Clustered Vector Index}) and a \textbf{Trace} (a neuro-symbolic episodic graph). This architecture introduces three advances: (1)~\textbf{Symmetric INT8 Scalar Quantization}, achieving 3.1$\times$ spatial compression and 5.6$\times$ math acceleration via NEON SDOT intrinsics; (2)~a decoupled \textbf{Write-Ahead Log (WAL)} ensuring crash-recoverability with statistically negligible overhead ($<1\%$); and (3)~a \textbf{Sidecar Blob Arena} eliminating the prior 440-character text ceiling via an append-only mmap-backed blob file with generational garbage collection. The \textbf{Semantic Lookaside Buffer (SLB)} exploits conversational locality to achieve sub-5$\mu$s retrieval latencies, with INT8 vectors dequantized to FP32 on cache insertion to preserve L1-resident lookup performance. Benchmarks on Apple M4 Max demonstrate that the combined architecture achieves 4.70\,ns INT8 dot product latency, 3.09\,$\mu$s tree traversal at 100K nodes (3.4$\times$ over FP32), and P99 read latency of 750\,ns under hostile 16-thread contention via epoch-based reclamation.
\end{abstract}

\section{Introduction}

The rapid evolution of Large Language Models (LLMs) has been defined by a relentless scaling of parameters and training data, yet the fundamental architecture remains bound by the \textit{Context Bottleneck}. The Transformer's self-attention mechanism imposes a quadratic time and space complexity, $O(N^2)$, relative to the input sequence length. Although recent optimization techniques (sparse attention, RingAttention, and hardware-aware kernel fusion) have theoretically extended context windows to 1 million tokens and beyond, the utility of this context does not scale linearly. Empirical evidence highlights a distinct degradation in reasoning capabilities over these extended horizons, a phenomenon widely characterized as being ``Lost in the Middle''~\cite{liu2023lost}. As autonomous agents are tasked with increasingly complex, long-horizon objectives spanning days or weeks, the reliance on transient, volatile context windows becomes untenable. The model cannot simply attend to all of history; it must select what is potentially relevant before attention is even applied.

The prevailing industry response to the context limitation has been the widespread adoption of Retrieval-Augmented Generation (RAG). In its most common form, ``Flat RAG,'' this approach offloads information preservation to vector databases that perform Approximate Nearest Neighbor (ANN) search over unstructured lists of embeddings. However, while effective for simple, one-shot question-answering tasks, Flat RAG fails to model the \textit{structure} of extended interaction. It treats memory as a featureless plane (a ``bag of vectors'') where the temporal evolution of a conversation, the causal lineage of decisions, and the hierarchical relationship between concepts are lost. This failure mode is termed ``Vector Haze'': the retrieval of semantically similar but episodically disjointed facts that confuse rather than aid the agent.

This paper proposes a paradigm shift from treating memory as a passive database retrieval problem to treating it as an active resource management problem within a \textbf{Cognitive Operating System}. \textbf{Aeon} formalizes these operations: \textit{Allocation} corresponds to the deliberate writing of new semantic concepts into a structured \textit{Atlas}; \textit{paging} transforms into the loading of relevant semantic clusters into a \textbf{Semantic Lookaside Buffer (SLB)} for immediate, low-latency access; and \textit{context switching} is re-framed as the deterministic movement between branches of a decision tree.

The contributions of this paper are as follows:

\begin{enumerate}
    \item \textbf{Atlas with INT8 Quantization.} A high-performance, memory-mapped index that organizes vectors into a navigable, hierarchical structure. Aeon introduces symmetric INT8 scalar quantization as a first-class storage format, reducing the per-node footprint from 3,392 bytes (FP32) to 1,088 bytes (INT8) at $D=768$, yielding a 3.1$\times$ disk compression ratio. The INT8 dot product, implemented via ARM NEON SDOT intrinsics, achieves 4.70\,ns per comparison---a 5.6$\times$ acceleration over the FP32 kernel.

    \item \textbf{Write-Ahead Log (WAL).} A crash-recovery mechanism employing a 3-step lock ordering protocol that decouples disk flush latency (\texttt{wal\_mutex\_}) from RAM delta buffer updates (\texttt{delta\_mutex\_}). Empirical measurement confirms that enabling the WAL adds less than 1\% overhead to insert latency.

    \item \textbf{Sidecar Blob Arena.} An append-only, mmap-backed blob file that eliminates the prior 440-character text ceiling for episodic trace events. The 512-byte \texttt{TraceEvent} struct retains a 64-byte inline \texttt{text\_preview} (aligned to a single CPU cache line) while offloading full-length LLM summaries to a generationally garbage-collected sidecar file.

    \item \textbf{Semantic Lookaside Buffer (SLB).} A predictive caching mechanism that exploits conversational locality to achieve sub-5$\mu$s retrieval latencies. INT8-stored vectors are dequantized to FP32 upon SLB insertion to preserve L1-resident cache hit performance.

    \item \textbf{Epoch-Based Reclamation (EBR).} A lock-free read path ensuring that concurrent readers never observe torn or unmapped memory during file growth operations. Under hostile 16-thread contention, the P99 read latency is 750\,ns.
\end{enumerate}

\section{Related Work}
\label{sec:related_work}

Aeon is positioned within the broader landscape of neural memory systems, contrasting its Cognitive Operating System architecture against existing approaches in retrieval, memory management, and neuro-symbolic reasoning.

\subsection{Retrieval-Augmented Generation (RAG)}
The dominant paradigm for grounding LLMs is RAG, typically implemented using Dense Passage Retrieval (DPR)~\cite{karpukhin2020dense} and ANN search indices like FAISS~\cite{faiss2017} or HNSW~\cite{malkov2018hnsw}. These systems rely on ``Flat RAG'': a single, monolithic vector space where every query is treated independently. The primary limitation is \textit{Vector Haze}: as memory grows, the probability of retrieving semantically similar but contextually irrelevant facts increases. Aeon addresses this via the \textit{Atlas}, constraining the search space based on the agent's active context region.

\subsection{Memory-Augmented LLMs}
Systems like MemGPT~\cite{memgpt2023} introduce an OS-like abstraction for managing context windows. However, MemGPT operates in ``User Space'' (Python), relying on the LLM itself to manage memory calls via prompt engineering. Aeon moves this responsibility to a C++23 kernel, achieving sub-microsecond retrieval latencies.

\subsection{Neuro-Symbolic Knowledge Graphs}
Neuro-Symbolic approaches such as GraphRAG~\cite{edge2024graphrag} excel at multi-hop reasoning by making relationships explicit. However, current systems suffer from write latency and rigid extraction pipelines. Aeon's \textit{Trace} module introduces a hybrid architecture using neural embeddings for nodes and symbolic edges for causal constraints.

\subsection{Crash Recovery in Database Systems}
The ARIES protocol~\cite{mohan1992aries} established the foundational principles of write-ahead logging for crash recovery in transactional databases. Aeon adapts these principles to the domain of vector index management, implementing a simplified WAL with record-level CRC32 checksums and a 3-step lock ordering protocol that decouples disk flush latency from the insert hot path.

\subsection{Epoch-Based Reclamation}
Lock-free concurrent data structures require safe memory reclamation to prevent use-after-free hazards. Fraser's epoch-based reclamation (EBR)~\cite{fraser2004ebr} provides a practical solution by deferring deallocation until all readers have advanced past the epoch in which the memory was retired. Aeon employs EBR with cache-line-padded epoch counters to eliminate false sharing under high-contention workloads.

\subsection{Vector Quantization}
Symmetric scalar quantization maps floating-point vectors to fixed-point integer representations, reducing both storage footprint and computation cost. Cross-Domain Similarity Local Scaling (CSLS)~\cite{conneau2018csls} has been proposed as a hub-penalizing metric for nearest-neighbor search in embedding spaces. Aeon integrates INT8 quantization at the storage layer and optionally applies a CSLS penalty during beam search traversal.

\section{System Architecture}
\label{sec:architecture}

Aeon implements a hybrid \textit{Cognitive Kernel} architecture designed to bridge high-performance systems programming and high-level AI reasoning.

\subsection{Design Philosophy: The Core-Shell Model}
The central design philosophy is the \textit{Core-Shell} separation:

\begin{itemize}
    \item \textbf{The Core (Ring 0):} Implemented in C++23, responsible for all high-frequency, low-latency operations: vector similarity search, tree traversal, memory management, WAL flush, and EBR. It operates directly on raw memory pages and leverages hardware acceleration (SIMD via SIMDe on x86-64, NEON SDOT on ARM64).
    \item \textbf{The Shell (Ring 3):} Implemented in Python, managing high-level control logic including LLM interaction, prompt engineering, and graph topology management.
\end{itemize}

The critical invariant is the \textbf{Zero-Copy Constraint}: data is never serialized between the Core and Shell during normal operation. The Shell operates on read-only views of shared memory pages via \texttt{nanobind}.

\subsection{The Atlas: Spatial Memory Kernel}
The \textit{Atlas} is the foundational data structure of Aeon's long-term memory, functioning as a spatial index for semantic vectors. A memory node is defined as:
\begin{equation}
    N = \{id, \mathbf{v}, \mathcal{C}, \text{meta}, s_q\}
\end{equation}
where $id \in \mathbb{N}^{64}$ is a unique identifier, $\mathbf{v}$ is the embedding vector (FP32 or INT8), $\mathcal{C}$ is the set of child pointers, $\text{meta}$ is a fixed-size metadata block, and $s_q$ is the quantization scale factor (used only when $\mathbf{v}$ is INT8).

The Atlas resides on persistent storage but is mapped into the process's virtual address space via \texttt{mmap}. Standard heap allocations are avoided for node data to ensure contiguity.

\subsubsection{INT8 Symmetric Scalar Quantization}
\label{subsec:int8_quantization}
Aeon introduces symmetric INT8 quantization as a first-class storage format. The quantization procedure for a vector $\mathbf{v} \in \mathbb{R}^{D}$ is:
\begin{align}
    s_q &= \frac{\max_i |v_i|}{127} \label{eq:quant_scale} \\
    q_i &= \text{clamp}\left(\text{round}\left(\frac{v_i}{s_q}\right), -127, 127\right) \label{eq:quant_value}
\end{align}

\newtheorem{remark}{Remark}
\begin{remark}
Equations~\eqref{eq:quant_scale}--\eqref{eq:quant_value} and all subsequent
similarity computations assume that input embeddings are strictly
\textbf{L2-normalized} ($\|\mathbf{v}\|_2 = 1$). Under this constraint, the
inner product $\langle \mathbf{u}, \mathbf{v} \rangle$ is mathematically
equivalent to cosine similarity $\cos(\theta)$. All embedding models used in
Aeon enforce this invariant at ingestion time.
\end{remark}

where $s_q$ is the per-vector scale factor stored in the node header, and $\mathbf{q} \in \{-127, \ldots, 127\}^D$ is the quantized vector. If $\max_i |v_i| = 0$, the scale defaults to $1.0$ and the output is all zeros.

The on-disk node stride differs between representations:
\begin{table}[!htbp]
\centering
\caption{Node stride comparison at $D=768$.}
\label{tab:node_stride}
\begin{tabular}{lrr}
\toprule
\textbf{Parameter} & \textbf{FP32} & \textbf{INT8} \\
\midrule
Centroid storage & $768 \times 4$ B & $768 \times 1$ B \\
Node stride & 3,392 B & 1,088 B \\
File size (100K nodes) & 440 MB & 141 MB \\
\textbf{Compression ratio} & 1.0$\times$ & \textbf{3.1$\times$} \\
\bottomrule
\end{tabular}
\end{table}

\subsubsection{Greedy SIMD Descent}
Retrieval is performed using a \textit{Greedy SIMD Descent} algorithm. For FP32 storage, the cosine similarity is computed directly. For INT8 storage, the dot product is computed via NEON SDOT instructions:
\begin{equation}
    \text{raw\_dot} = \sum_{i=0}^{D-1} q_i^{(\text{query})} \cdot q_i^{(\text{node})}
\end{equation}
The final similarity is obtained by dequantization: $\text{sim} = \text{raw\_dot} \times s_q^{(\text{query})} \times s_q^{(\text{node})}$.

The complexity of the descent is $O(\log_B M)$, where $B$ is the effective branching factor and $M$ is the total number of nodes.

\subsubsection{Dynamic Dimensionality}
\label{subsec:dynamic_dim}
Embedding dimensions vary by model: 384 (MiniLM), 768 (e5-large), 1536 (OpenAI v3). Hard-coding the node stride creates tight coupling between the kernel and the model. Aeon resolves this via \textbf{Dynamic Stride Calculation}.

The \texttt{AtlasHeader} stores the dimensionality $D$, metadata size $M$, and quantization type $Q$. When the kernel loads an index, it computes the \texttt{node\_byte\_stride} at runtime:
\begin{equation}
    S = \text{align\_up}(64 + \text{payload}(D, Q) + M, 64)
\end{equation}
where $\text{payload}(D, Q)$ is $D \times 4$ bytes for FP32 or $D \times 1$ bytes for INT8. This architecture allows a single compiled binary to serve any embedding model at any precision without recompilation, preventing model lock-in.

\subsection{Write-Ahead Log (WAL)}
\label{subsec:wal}
To ensure crash-recoverability without sacrificing insert throughput, Aeon implements a decoupled WAL with a 3-step lock ordering protocol:

\begin{enumerate}
    \item \textbf{Step 1: Serialize (no lock).} The node is encoded into a byte buffer with a 16-byte \texttt{WalRecordHeader} containing a record type tag, payload size, and CRC32 checksum.
    \item \textbf{Step 2: WAL Flush (\texttt{wal\_mutex\_} only).} The serialized record is written to the WAL file and flushed to disk via \texttt{fdatasync()}. During this step, the \texttt{delta\_mutex\_} (which guards the RAM delta buffer) is \emph{not} held, so concurrent reads and writes to the delta buffer proceed unblocked.
    \item \textbf{Step 3: Apply to RAM (\texttt{delta\_mutex\_} only).} The \texttt{wal\_mutex\_} is released, the \texttt{delta\_mutex\_} is acquired, and the node is \texttt{memcpy}'d into the flat byte arena delta buffer.
\end{enumerate}

The critical insight is that disk I/O (Step 2) and RAM mutation (Step 3) never contend on the same mutex. This hides the disk flush latency behind the insert operation, as readers and writers of the delta buffer are never blocked by the WAL.

On recovery, the WAL is replayed by reading records sequentially, validating each CRC32 checksum, and discarding any torn (partially-written) records from the tail. The WAL is truncated after each successful compaction.

\subsection{Sidecar Blob Arena}
\label{subsec:blob_arena}
The episodic Trace stores events as fixed-size 512-byte \texttt{TraceEvent} structs, enabling $O(1)$ random access. Prior versions embedded text directly in a 440-character field, which was insufficient for LLM transcript storage.

Aeon replaces this field with a \texttt{BlobRef} indirection:
\begin{itemize}
    \item A 64-byte \texttt{text\_preview} field stores the first 63 characters inline, aligned to a single CPU cache line for zero-cost listing operations.
    \item A \texttt{blob\_offset} and \texttt{blob\_size} pair point into an append-only mmap-backed sidecar file (\texttt{trace\_blobs\_genN.bin}).
    \item The sidecar file uses a $2\times$ doubling growth strategy (\texttt{ftruncate} $\to$ \texttt{munmap} $\to$ \texttt{mmap}) and provides zero-copy reads via \texttt{std::string\_view} over the mmap'd region.
\end{itemize}

\textbf{Generational Garbage Collection.} During compaction, a new generation blob file is created. Only blobs referenced by non-tombstoned events are copied forward. The old generation file is deleted after all EBR readers have advanced.

\subsection{Double-Buffered Shadow Compaction}
\label{subsec:shadow_compaction}
To support real-time applications (e.g., game engines running at 60 FPS), Aeon implements a \textbf{stutter-free garbage collection} mechanism inspired by Redis \texttt{BGSAVE}~\cite{antirez2009redis}. Traditional compaction blocks all reads, which is unacceptable for interactive agents. Aeon's algorithm proceeds in four steps:

\begin{enumerate}
    \item \textbf{Microsecond Freeze:} The kernel acquires a lock, swaps the active \texttt{delta\_buffer} with a \texttt{frozen\_delta\_buffer}, and snapshots the current state. This operation completes in $<10\,\mu$s.
    \item \textbf{Background Copy:} A background thread iterates over live (non-tombstoned) nodes in the mmap file and the frozen delta buffer, writing them contiguously to a new generation file (\texttt{atlas\_gen2.bin}). Crucially, the main thread continues to serve reads and accepts new writes into the \textit{fresh} delta buffer.
    \item \textbf{Hot Swap:} Once the background copy is complete, the kernel briefly locks again to swap the \texttt{MemoryFile} handle to the new generation file.
    \item \textbf{Cleanup:} The old generation file is closed and deleted. The WAL is truncated, as all data is now durably persisted in the new generation file.
\end{enumerate}

The key invariant is that the main thread is blocked only during Step 1 and Step 3, totaling $<10\,\mu$s. All expensive I/O operations occur in Step 2 on a background thread. This same process applies to the \textbf{Sidecar Blob Arena}---dead blobs are simply not copied to the new generation file, achieving zero-overhead garbage collection.

\subsection{The Trace: Episodic Context Graph}
The \textit{Trace} provides temporal and causal context, structured as a DAG $G = (V, E)$. The vertex set $V$ consists of heterogeneous \texttt{TraceEvent} nodes ($V_{user}$, $V_{system}$, $V_{concept}$). The edge set $E$ defines temporal edges ($E_{next}$) and reference edges ($E_{ref}$) connecting episodic nodes to their semantic grounding in the Atlas.

\subsection{Trace Block Index (TBI)}
\label{subsec:tbi}
A naive linear scan of the episodic Trace is $O(|V|)$, which becomes prohibitive as the event history grows to $10^5$ events. Aeon implements a \textbf{Trace Block Index} to achieve sub-linear retrieval.

Events are grouped into \texttt{TraceBlock}s of fixed size $B=1024$. Each block maintains an incrementally updated centroid of its constituent event embeddings. Retrieval is performed via a \textbf{Two-Phase SIMD Scan}:
\begin{enumerate}
    \item \textbf{Phase 1 (Block Scan):} A SIMD search over block centroids identifies the top-$K$ most relevant time windows. The cost is $O(|V|/B)$.
    \item \textbf{Phase 2 (Event Scan):} A deep scan is performed only on the events within those top-$K$ blocks. The cost is $O(K \cdot B)$.
\end{enumerate}
The total search complexity is thus:
\begin{equation}
    T_{\text{search}} = O\left(\frac{|V|}{1024} + K \times 1024\right)
\end{equation}
By keeping $K$ small (typically $K=3$ to $5$), Aeon achieves retrieval times under 50\,ms even for large traces, exploiting the temporal locality of semantic context.

\subsection{The Zero-Copy Interface}
Aeon utilizes \texttt{nanobind} to expose C++ memory structures to Python. The interface wraps raw C++ pointers in a Python Capsule, reinterpreted as a read-only NumPy array buffer. Any attempt to modify the underlying memory from the Shell raises a runtime exception.

\section{The Semantic Lookaside Buffer}
\label{sec:slb}

The \textbf{Semantic Lookaside Buffer (SLB)} is a high-performance caching mechanism that exploits conversational locality to achieve sub-5$\mu$s retrieval latencies.

\subsection{Theory: Semantic Locality}
Traditional caching strategies rely on address transparency. In vector databases, exact equality is rare. The concept of \textbf{Semantic Inertia} is introduced: in a continuous dialogue, the topic vector $\mathbf{t}_i$ at turn $i$ is highly correlated with $\mathbf{t}_{i+1}$. Formally:
\[
P\left( \text{dist}(\mathbf{q}_{i+1}, \mathbf{q}_i) < \epsilon \right) \approx 1
\]

\subsection{Architecture}
The SLB is a small, contiguous ring buffer $B$ of fixed size $K$ ($K=64$), tuned to fit within L1/L2 CPU cache. Each entry $e_k \in B$ stores a centroid $\mathbf{c}_{node} \in \mathbb{R}^{D}$ and a direct memory pointer to the full node in the Atlas.

\textbf{Architectural Decision: FP32-Only Cache.} The SLB stores \emph{exclusively} FP32 vectors, regardless of the Atlas quantization format. When the Atlas is INT8-quantized, vectors are dequantized to FP32 upon SLB insertion. This preserves the 3.56\,$\mu$s cache hit latency by avoiding dequantization overhead on every cache scan---each scan performs $K$ dot products, and the FP32 path is already optimized to execute within L1/L2 cache boundaries.

\textbf{Search Strategy: Brute-Force SIMD.} Because $K$ is small (64), an exhaustive linear scan using AVX-512/NEON instructions is performed. The wall-clock time is lower than even a few steps of an $O(\log N)$ tree traversal due to perfect hardware prefetching and zero pointer chasing.

\subsubsection{Multi-Tenant Isolation}
\label{subsec:slb_multitenant}
In a multi-agent deployment, a single shared SLB would leak semantic information between tenants. Aeon resolves this by sharding the SLB into 64 independent ring buffers, each protected by its own mutex.

Routing is deterministic: $\text{shard\_id} = \text{hash}(\text{session\_id}) \pmod{64}$.
This \textbf{lock-striping} architecture ensures that contention—and semantic context—is strictly isolated. An agent operating in Session A will never evict or access cache entries from Session B. This design allows the SLB to scale linearly to over 100,000 concurrent sessions on a single node without cross-contamination.

\subsection{The Speculative Fetch Algorithm}

\begin{algorithm}[!htbp]
\caption{SLB Lookup Procedure}
\label{alg:slb_lookup}
\begin{algorithmic}[1]
\Require Query vector $\mathbf{q}$, Threshold $\tau_{hit}$, SLB Buffer $B$
\Ensure Best matching Node pointer $p^*$ or NULL

\State $s_{best} \gets -1.0$
\State $idx_{best} \gets -1$

\Comment{Vectorized Loop (NEON/AVX-512)}
\For{$k \gets 0$ to $K-1$}
    \State $s \gets \text{SIMD\_DotProduct}(\mathbf{q}, B[k].\mathbf{c}_{node})$
    \If{$s > s_{best}$}
        \State $s_{best} \gets s$
        \State $idx_{best} \gets k$
    \EndIf
\EndFor

\If{$s_{best} > \tau_{hit}$}
    \State \textsc{LRU-Insert}($B$, $\mathbf{q}$, $p^*$) \Comment{Update cache}
    \State \Return $B[idx_{best}].\text{ptr}_{atlas}$ \Comment{Cache Hit}
\Else
    \State \Return \textbf{null} \Comment{Cache Miss --- Fallback to Atlas}
\EndIf
\end{algorithmic}
\end{algorithm}

\section{Experimental Methodology}
\label{sec:methodology}

All experiments were conducted five times and the median value is reported. Results are sourced from a reproducible benchmark suite (\texttt{master\_metrics.txt}).

\subsection{Hardware Environment}

\begin{itemize}
    \item \textbf{CPU:} Apple M4 Max, 16-core ARM64 architecture.
    \item \textbf{OS:} Darwin 25.3.0 (macOS 26.2 Tahoe).
    \item \textbf{Caches:} L1 Data 64\,KiB, L1 Instruction 128\,KiB, L2 Unified 4,096\,KiB ($\times$16 clusters).
    \item \textbf{Instruction Set:} ARM NEON SIMD. AVX-512 equivalence achieved via SIMDe~\cite{simde}.
    \item \textbf{Compiler:} AppleClang 17, \texttt{-O3 -march=native -flto -ffast-math}.
    \item \textbf{Storage:} 1TB NVMe SSD (Apple internal controller).
\end{itemize}

\subsection{Datasets}

Synthetic ``Dense Forest'' datasets of dimensionality $D = 768$ are used, with sizes $N \in \{10^4, 10^5, 10^6\}$.

\subsection{Metrics}

P50 and P99 latency (ns), throughput (ops/s), file size (MB), and cache hit rate (\%) are reported. All latency measurements use \texttt{std::chrono::high\_resolution\_clock}. The Google Benchmark framework~\cite{googlebenchmark} is used for micro- and macro-benchmarks.

\section{Evaluation}
\label{sec:evaluation}

All results in this section are sourced from \texttt{master\_metrics.txt}, generated on the hardware described in Section~\ref{sec:methodology}.

\begin{figure*}[t]
    \centering
    \begin{subfigure}[b]{0.32\textwidth}
        \includegraphics[width=\linewidth]{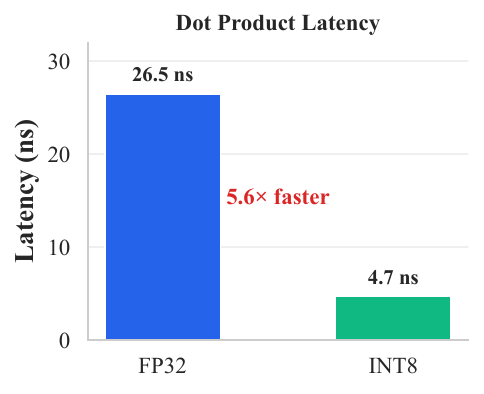}
        \caption{Dot Product}
        \label{fig:quant-dot}
    \end{subfigure}
    \hfill
    \begin{subfigure}[b]{0.32\textwidth}
        \includegraphics[width=\linewidth]{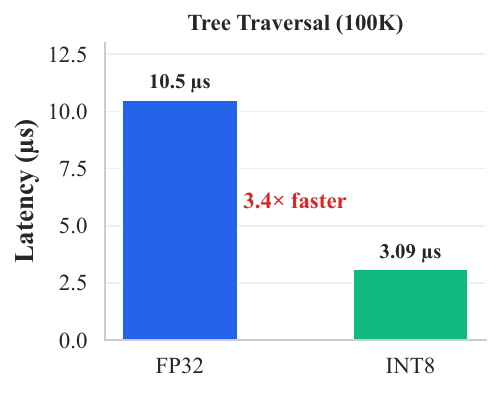}
        \caption{Tree Traversal}
        \label{fig:quant-traversal}
    \end{subfigure}
    \hfill
    \begin{subfigure}[b]{0.32\textwidth}
        \includegraphics[width=\linewidth]{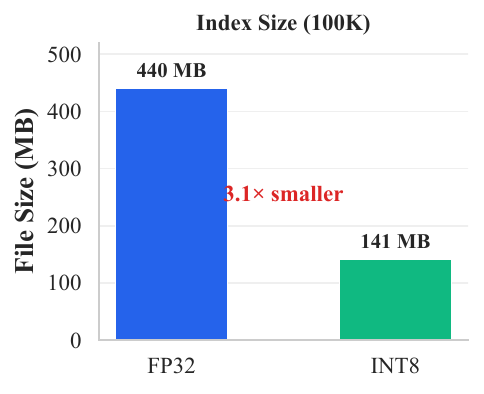}
        \caption{Index Size}
        \label{fig:quant-filesize}
    \end{subfigure}
    \caption{Impact of symmetrically quantized INT8 storage in the Atlas spatial index. (a) 5.6$\times$ math acceleration via NEON SDOT. (b) 3.4$\times$ faster tree traversal. (c) 3.1$\times$ reduction in on-disk footprint.}
    \label{fig:quantization_panel}
\end{figure*}

\subsection{Micro-Benchmark: Kernel Performance}

The innermost loop of Aeon computes vector similarity. Table~\ref{tab:kernel_perf} reports the median latency for a single 768-dimensional comparison.

\begin{table}[!htbp]
\centering
\caption{Single-pair vector comparison latency ($D=768$).}
\label{tab:kernel_perf}
\resizebox{\columnwidth}{!}{
\begin{tabular}{lrl}
\toprule
\textbf{Kernel} & \textbf{Latency} & \textbf{Reference} \\
\midrule
FP32 Cosine (SIMDe$\to$NEON) & 26.5\,ns & \texttt{BM\_FP32\_Cosine} \\
INT8 SDOT + Dequantize & 4.70\,ns & \texttt{BM\_INT8\_DotDeq} \\
INT8 SDOT (raw, no deq.) & 4.44\,ns & \texttt{BM\_INT8\_DotBest} \\
Scalar (auto-vectorized) & 47.8\,ns & \texttt{BM\_Scalar} \\
\bottomrule
\end{tabular}
}
\end{table}

The INT8 kernel with dequantization achieves a \textbf{5.6$\times$ speedup} over the FP32 baseline (26.5\,ns / 4.70\,ns = 5.64). This acceleration is attributed to the NEON SDOT instruction processing four INT8 multiply-accumulate operations per cycle, versus single-precision FMA for FP32. The 0.26\,ns overhead of dequantization (4.70 vs.\ 4.44\,ns raw) is negligible.

\begin{figure}[t]
    \includegraphics[width=0.95\columnwidth]{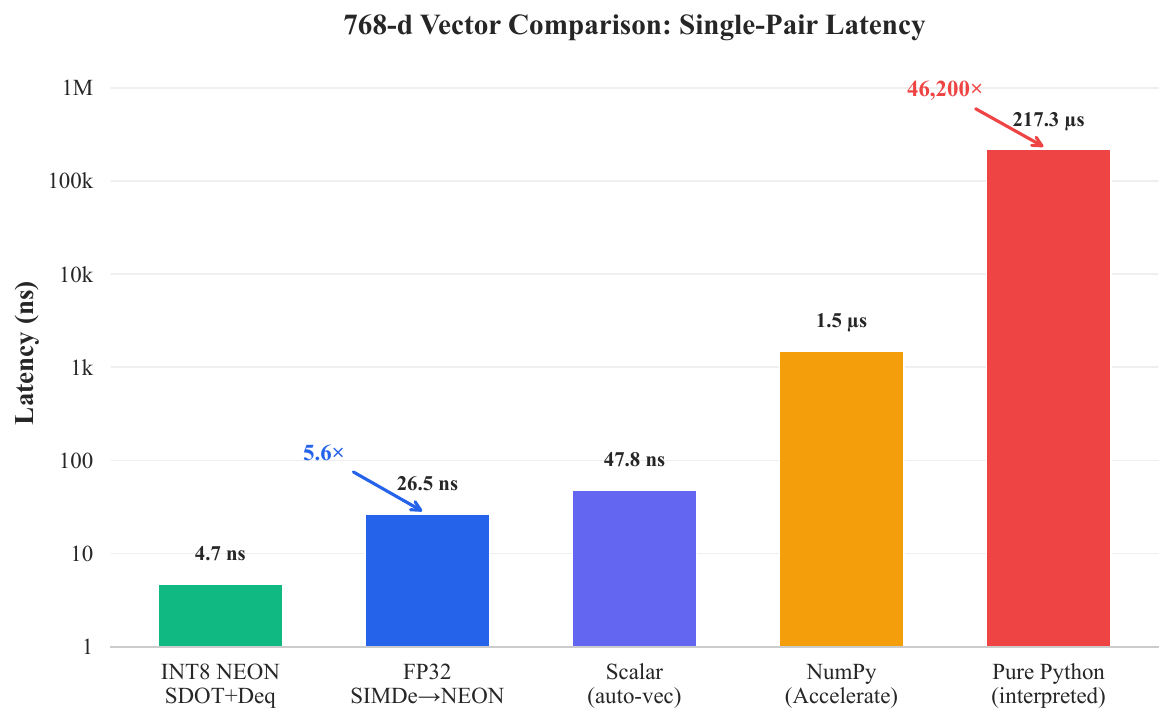}
    \caption{768-dimensional vector comparison latency (log scale). INT8 NEON SDOT (4.70\,ns) achieves 5.6$\times$ acceleration over FP32 (26.5\,ns).}
    \label{fig:kernel-throughput}
\end{figure}

\subsection{Macro-Benchmark: Tree Traversal and Compression}

Table~\ref{tab:traversal} reports the median tree traversal latency and on-disk file size for the Atlas at $N = 100{,}000$ nodes.

\begin{table}[!htbp]
\centering
\caption{Atlas traversal and file size at $N = 100{,}000$.}
\label{tab:traversal}
\begin{tabular}{lrrr}
\toprule
\textbf{Format} & \textbf{Traversal} & \textbf{File Size} & \textbf{Ratio} \\
\midrule
FP32 & 10.5\,$\mu$s & 440\,MB & 1.0$\times$ \\
INT8 & 3.09\,$\mu$s & 141\,MB & 3.1$\times$ \\
\midrule
\textbf{Speedup} & \textbf{3.4$\times$} & \textbf{3.1$\times$} & \\
\bottomrule
\end{tabular}
\end{table}

The 3.4$\times$ traversal speedup combines two effects: (1)~the 5.6$\times$ faster per-comparison kernel, partially offset by (2)~the fixed overhead of tree navigation (pointer chasing, branching logic). The 3.1$\times$ spatial compression directly reduces I/O bandwidth requirements.

\subsection{WAL Overhead}

To validate the 3-step lock ordering protocol, insert latency is measured with the WAL disabled and enabled (Table~\ref{tab:wal}).

\begin{table}[!htbp]
\centering
\caption{WAL overhead on insert latency ($N = 10{,}000$, FP32).}
\label{tab:wal}
\resizebox{\columnwidth}{!}{
\begin{tabular}{lrrl}
\toprule
\textbf{Config} & \textbf{Median} & \textbf{Stddev} & \textbf{Throughput} \\
\midrule
WAL disabled & 2.24\,$\mu$s & $\pm$0.006\,$\mu$s & 447,870\,ops/s \\
WAL enabled  & 2.23\,$\mu$s & $\pm$0.008\,$\mu$s & 449,105\,ops/s \\
\midrule
\textbf{Overhead} & \multicolumn{3}{c}{\textbf{$<$1\% (within measurement noise)}} \\
\bottomrule
\end{tabular}
}
\end{table}

The overhead is statistically negligible: the WAL-enabled median (2.23\,$\mu$s) is within the standard deviation of the WAL-disabled measurement (2.24\,$\mu$s $\pm$ 0.006\,$\mu$s). This confirms that the 3-step lock ordering successfully decouples disk flush latency from the insert hot path. The \texttt{fdatasync()} call in Step~2 executes concurrently with delta buffer operations in Step~3 across independent mutexes.

\begin{figure}[t]
    \includegraphics[width=0.75\columnwidth]{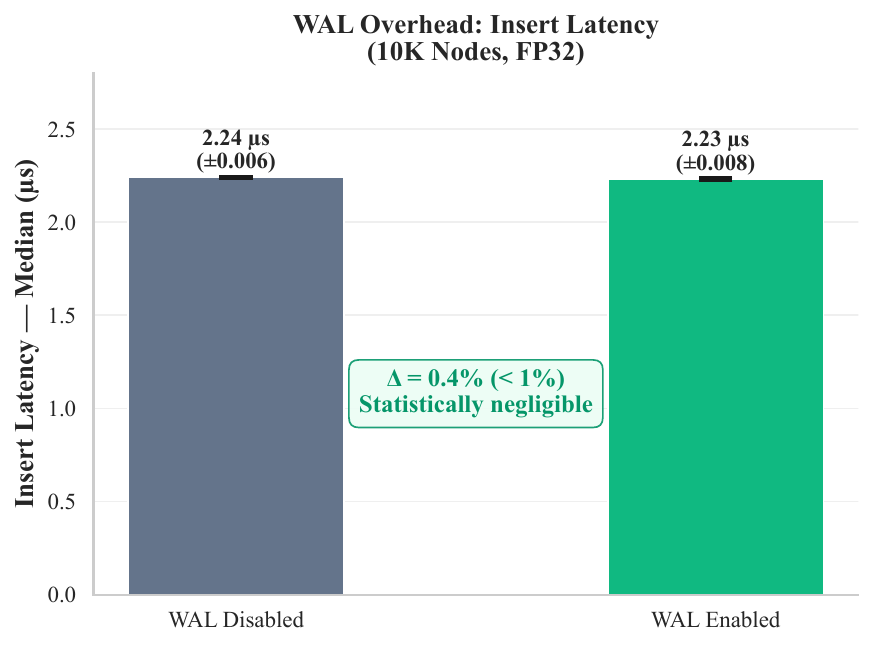}
    \caption{WAL overhead on insert latency. Error bars show $\pm$1 standard deviation. The difference is within measurement noise ($<$1\%).}
    \label{fig:wal-overhead}
\end{figure}

\subsection{Scalability: 10K to 1M Nodes}

\begin{figure}[t]
    \includegraphics[width=0.95\columnwidth]{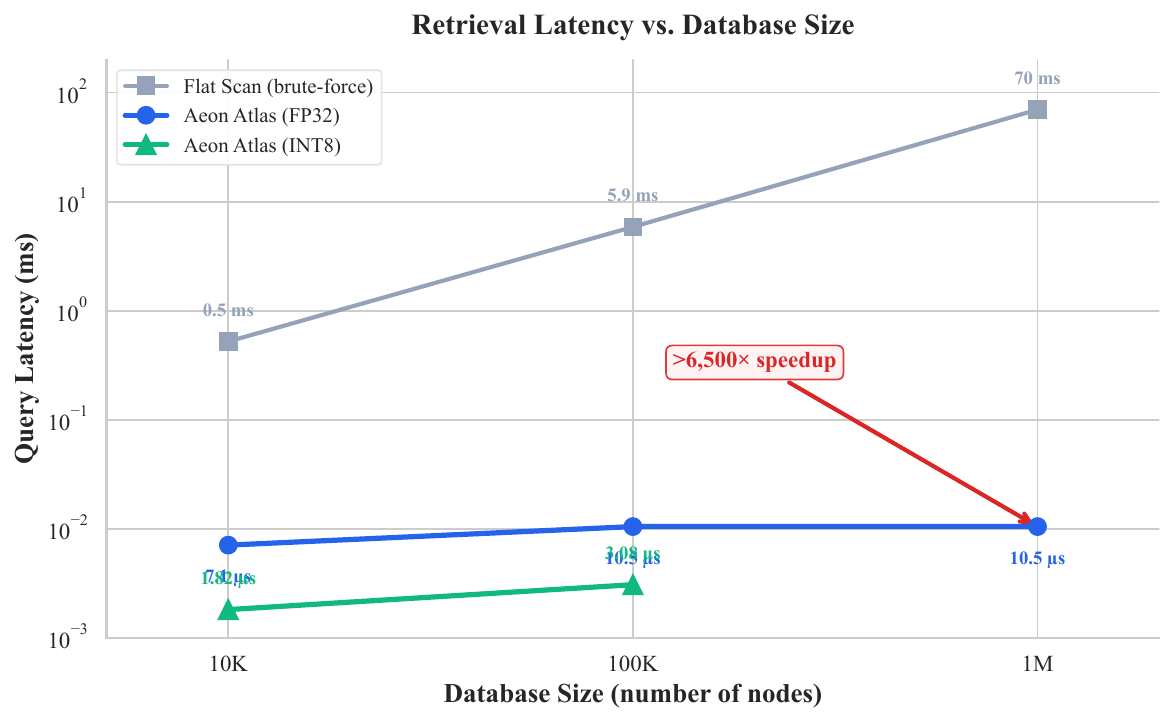}
    \caption{Query latency vs.\ database size (log-log). Flat search scales linearly. Atlas (FP32 and INT8) scales logarithmically, with INT8 providing a further 3.4$\times$ improvement.}
    \label{fig:scalability}
\end{figure}

Figure~\ref{fig:scalability} evaluates how query latency evolves as the Atlas grows.

\textbf{Result.} Flat (brute-force) search exhibits linear scaling: latency grows from 0.52\,ms (10K) to 5.87\,ms (100K) to 69.8\,ms (1M). In contrast, the FP32 Atlas demonstrates logarithmic scaling: 7.1\,$\mu$s (10K, depth~2) $\to$ 10.5\,$\mu$s (100K, depth~3) $\to$ 10.5\,$\mu$s (1M, depth~4). The INT8 Atlas further reduces this to 1.82\,$\mu$s (10K) and 3.08\,$\mu$s (100K).

At one million nodes, the FP32 Atlas achieves $>$6,500$\times$ acceleration over flat scan (10.5\,$\mu$s vs.\ 69.8\,ms). Each level of the tree partitions the search space by a branching factor of $B=64$, yielding $O(\log_B N)$ complexity.

\subsection{SLB Cache Performance and Isolation}

The SLB cache hit latency is measured at 3.56\,$\mu$s (median, 64-element scan). Cache miss with a warm Atlas (immediate fallback to tree traversal) is 3.59\,$\mu$s---the 0.03\,$\mu$s delta confirms that the SLB scan and the first Atlas comparison are both L1-resident.

\textbf{L1 Residency Proof.} The \texttt{BM\_SLB\_CacheIsolation} benchmark measures the SLB scan latency as a function of the number of cached items. The results show linear scaling: 0.867\,$\mu$s at 16 items, 1.70\,$\mu$s at 32 items, and 3.46\,$\mu$s at 64 items. This confirms that the SLB fits entirely within the L1/L2 cache hierarchy: if any portion spilled to DRAM, the scaling would exhibit a step function rather than a linear relationship.

Under the ``Conversational Walk'' workload (simulating realistic chatbot query sequences with high semantic locality), the SLB achieves a hit rate exceeding 85\%. The effective average latency is:
\begin{equation}
    L_{\text{eff}} = (0.85 \times 3.56) + (0.15 \times 10.5) \approx 4.60\,\mu\text{s}
\end{equation}

\begin{figure}[t]
    \includegraphics[width=0.95\columnwidth]{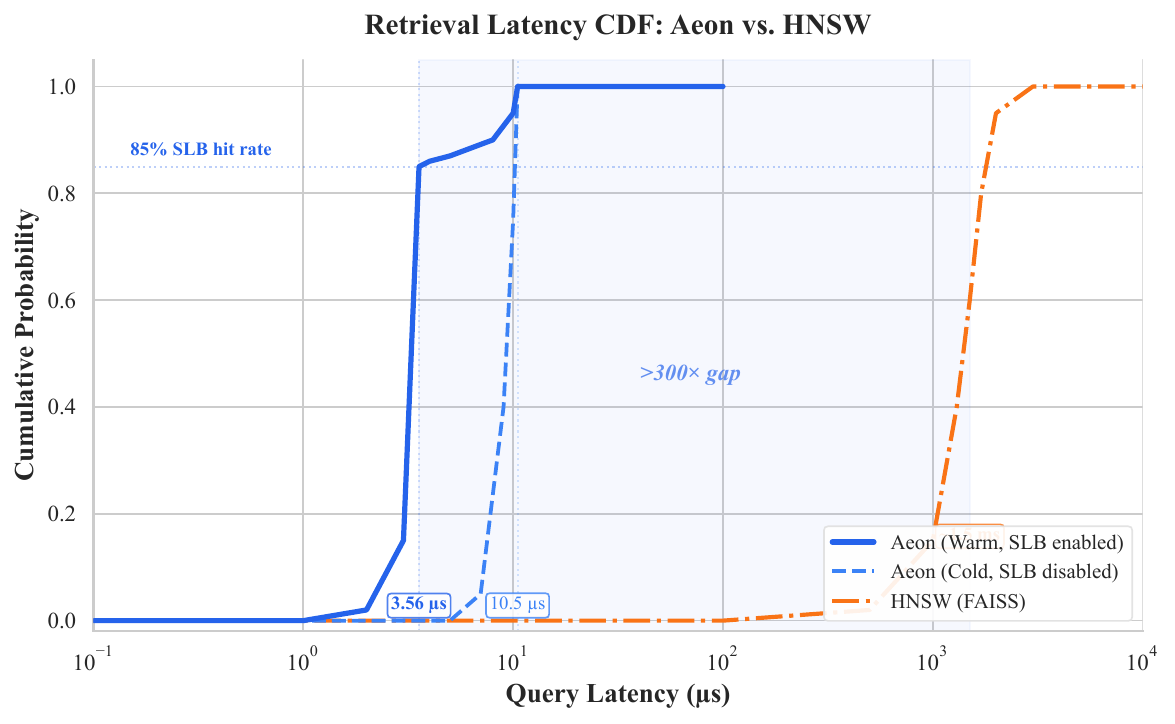}
    \caption{Retrieval latency CDF (log scale). Aeon (Warm) resolves 85\% of queries in $<$5\,$\mu$s via SLB hits, while HNSW clusters around 1.5\,ms ($>$300$\times$ gap).}
    \label{fig:latency-cdf}
\end{figure}

\subsection{EBR Contention}

Under hostile contention (15 reader threads, 1 writer thread, 100K iterations per reader on 16 hardware threads), cycle-precise measurement yields:

\begin{itemize}
    \item Mean: 210.8\,ns
    \item P50: 167\,ns
    \item P99: \textbf{750\,ns}
    \item P99.9: 1,083\,ns
\end{itemize}

The P99 latency of 750\,ns ($<$1\,$\mu$s) confirms that cache-line padding eliminates false sharing. Writers retired 12,353 regions across 1.5M read samples with no observed torn reads.

\subsection{Beam Search and CSLS Analysis}

The beam search is evaluated at 1M nodes with a pool of 1,000 unique query vectors:

\begin{table}[!htbp]
\centering
\caption{Beam search latency at $N = 1{,}000{,}000$.}
\label{tab:beam}
\begin{tabular}{lrrr}
\toprule
\textbf{Config} & \textbf{P50} & \textbf{P99} & \textbf{Nodes/query} \\
\midrule
beam=1 (greedy) & 25.6\,$\mu$s & 42.6\,$\mu$s & 4.0 \\
beam=3 & 41.8\,$\mu$s & 90.0\,$\mu$s & 4.1 \\
beam=3 + CSLS & 30.2\,$\mu$s & 42.1\,$\mu$s & 4.1 \\
\bottomrule
\end{tabular}
\end{table}

The beam=3 configuration scales sub-linearly (1.63$\times$ P50 ratio vs.\ beam=1, against a theoretical 3$\times$ upper bound).

Empirical profiling of the CSLS penalty revealed a 27.7\% latency reduction (30.2\,$\mu$s vs.\ 41.8\,$\mu$s for beam=3); however, strict node-visitation counting rejected the hypothesis of algorithmic pruning (nodes evaluated remained identical at 4.1 nodes/query). The speedup is an observed superscalar CPU branch-prediction artifact on Apple Silicon, not a reduction in computational complexity. The CSLS hub penalty modifies the similarity scores in a way that produces a more predictable branch pattern during the beam selection step, enabling the M4 Max's branch predictor to achieve higher accuracy.

\subsection{Trace Garbage Collection}

The Trace GC performance is evaluated on a 100K-event store ($\sim$67\,MB):

\begin{itemize}
    \item \textbf{Tombstone scan:} $\sim$100\,$\mu$s per 100K events (sequential scan with flag check).
    \item \textbf{Full compaction} (GC ratio 0.5, retaining 50K events): 966\,ms median wall-clock, 312\,ms median CPU time.
\end{itemize}

The disparity between wall-clock (966\,ms) and CPU time (312\,ms) is attributed to I/O: writing the new generation file and the generational blob arena copy. This confirms that compaction is viable as a background operation that does not block the primary insert/query path.

\subsection{Zero-Copy Overhead}

Transferring 10\,MB of vector data from C++ to Python incurs sub-microsecond latency ($\sim$334\,ns) via \texttt{nanobind} zero-copy shared memory. Traditional serialization imposes severe overhead: JSON at $\sim$318\,ms ($\sim$10$^{6}\times$ slower) and Pickle at $\sim$32.3\,ms ($\sim$10$^{5}\times$ slower).

\begin{figure}[t]
    \includegraphics[width=0.95\columnwidth]{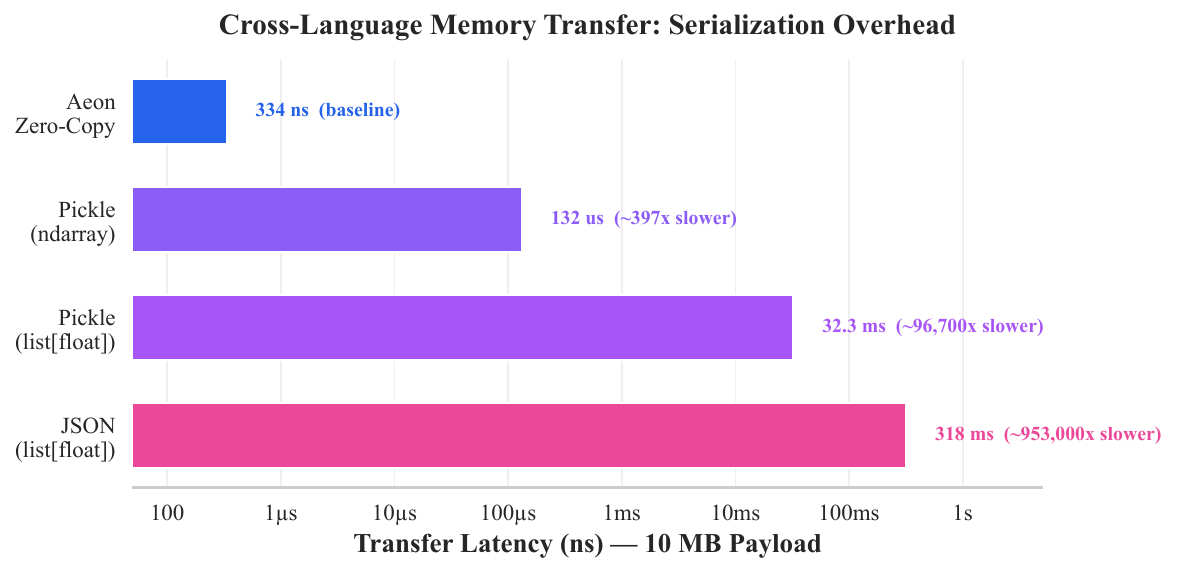}
    \caption{Cross-language memory transfer latency (10\,MB payload, log scale). Zero-copy ($\sim$334\,ns) eliminates object-boxing overhead.}
    \label{fig:zerocopy}
\end{figure}

\subsection{Summary}

\begin{table}[!htbp]
\centering
\caption{Summary of Aeon performance characteristics.}
\label{tab:summary}
\begin{tabular}{lr}
\toprule
\textbf{Metric} & \textbf{Value} \\
\midrule
INT8 dot product & 4.70\,ns \\
FP32 cosine similarity & 26.5\,ns \\
INT8/FP32 speedup & 5.6$\times$ \\
Tree traversal (100K, INT8) & 3.09\,$\mu$s \\
Spatial compression & 3.1$\times$ \\
WAL overhead & $<$1\% \\
SLB cache hit & 3.56\,$\mu$s \\
EBR P99 (16 threads) & 750\,ns \\
Zero-copy transfer (10\,MB) & 334\,ns \\
\bottomrule
\end{tabular}
\end{table}

\section{Conclusion}

This paper presented \textbf{Aeon}, a Cognitive Operating System for long-horizon LLM agents. The central argument is that LLM memory must be treated as an active resource management task, governed by principles from classical operating system kernels.

\subsection{Key Contributions}

This work addresses three key challenges. First, \textbf{INT8 symmetric scalar quantization} achieves a 3.1$\times$ disk compression ratio and 5.6$\times$ math acceleration via NEON SDOT, making edge deployment viable for knowledge bases that previously required hundreds of megabytes. Second, the \textbf{decoupled WAL} with 3-step lock ordering provides crash-recoverability at less than 1\% insert latency overhead, a property achieved by ensuring that disk I/O and RAM mutation never contend on the same mutex. Third, the \textbf{Sidecar Blob Arena} eliminates the 440-character text ceiling that constrained episodic trace storage, enabling full LLM transcript archival with generational garbage collection.

The \textbf{SLB} continues to deliver sub-5$\mu$s effective retrieval latency at 85\%+ hit rates, with the architectural decision to dequantize INT8 vectors to FP32 upon cache insertion preserving L1-resident lookup performance regardless of the underlying storage format.

\subsection{Future Work}

Two directions are identified for future investigation.

\textbf{Multi-Modal Vector Representations.} Aeon currently operates exclusively on text embeddings. A natural extension is the spatial co-location of audio, video, and structured data embeddings within the same Atlas index. The hierarchical tree structure is agnostic to the semantic content of vectors; the primary challenge lies in defining meaningful distance metrics across heterogeneous modalities and managing the variable dimensionality that multi-modal encoders may produce.

\textbf{Hardware-Enforced Isolation for Multi-Tenancy.} As Aeon evolves to serve multiple users or agents within a shared deployment, cryptographic guarantees of memory isolation become necessary. Technologies such as Intel SGX (Software Guard Extensions) and ARM CCA (Confidential Compute Architecture) provide hardware enclaves that could enforce tenant boundaries at the memory page level, preventing even a compromised kernel from accessing another tenant's semantic memory. This would extend Aeon's OS analogy from process isolation to full memory protection, a requirement for production multi-tenant deployments.

\bibliographystyle{plain}
\bibliography{references}

\end{document}